\documentclass[journal]{IEEEtran}

\usepackage{amsfonts}
\usepackage{amsmath}
\usepackage{amssymb}
\usepackage{mathabx}
\usepackage{mathrsfs}
\usepackage{tensor}
\usepackage{esint}
\usepackage{tikz}
\usepackage{multirow}
\usepackage{makecell}
\usepackage{textcomp}
\usepackage{theorem}
\usepackage{epstopdf}
\usepackage[normalem]{ulem}
\usepackage{booktabs,floatrow}
\newtheorem{theorem}{Theorem}[section]

\newtheorem{corollary}{Corollary}[section]

\newcolumntype{C}[1]{>{\centering\arraybackslash}m{#1}}

\DeclareMathAlphabet{\mathpzc}{OT1}{pzc}{m}{it}
\DeclareMathOperator*{\argmin}{argmin}

\newcommand{\x}{\mbox{\textbf{x}}}
\newcommand{\y}{\mbox{\textbf{y}}}

                    {$\blacksquare$\vspace*{7pt}} 

\def\f{\frac}

\def\bi{{\mathbf i}}

\def\n{\mathbf{n}}

\def\x{\boldsymbol{x}}
\def\y{{\boldsymbol y}}

\def\I{{\mathbf I}}

\def\X{{\mathbf X}}
\def\Y{{\mathbf Y}}

\def\mT{{\mathcal T}}

\def\bi{\begin{itemize}} \def\ei{\end{itemize}}
\def\be{\begin{eqnarray*}}
\def\ee{\end{eqnarray*}}

\def\0{{\mathbf 0}}

\newcommand{\beq}{\begin{equation}}
\newcommand{\eeq}{\end{equation}}

\def\eref#1{(\ref{#1})}

\def\x{\mathbf{x}}
\def\y{\mathbf{y}}

\def\XXint#1#2#3{{\setbox0=\hbox{$#1{#2#3}{\int}$ }
\vcenter{\hbox{$#2#3$ }}\kern-.55\wd0}}

\begin{document}

\title{Automatic 3D Registration of Dental CBCT and Face Scan Data using 2D Projection Images}
	
\author{\IEEEauthorblockN{Hyoung Suk Park\IEEEauthorrefmark{1}, Chang Min Hyun\IEEEauthorrefmark{2}, Sang-Hwy Lee\IEEEauthorrefmark{3}, Jin Keun Seo\IEEEauthorrefmark{2} and Kiwan Jeon\IEEEauthorrefmark{1}}
	\\ \IEEEauthorblockA{\IEEEauthorrefmark{1}National Institute for Mathematical Sciences, Daejeon, 34047, Republic of Korea} 
	\\ \IEEEauthorblockA{\IEEEauthorrefmark{2}School of Mathematics and Computing (Computational Science and Engineering), Yonsei University, Seoul, 03722, Republic of Korea}
	\\ \IEEEauthorblockA{\IEEEauthorrefmark{3}Department of Oral and Maxillofacial Surgery, Oral Science Research Center, College of Dentistry, Yonsei University, Seoul, 03722, Republic of Korea}
	\thanks{Manuscript received XXX; revised  XXX. Corresponding author: Kiwan Jeon (jeonkiwan@nims.re.kr).}}
	
\markboth{}%
	{ \MakeLowercase{\textit{Park et al.}}: }
	
\IEEEtitleabstractindextext{%

\begin{abstract}
This paper presents a fully automatic registration method of dental cone-beam computed tomography (CBCT) and face scan data. It can be used for a digital platform of 3D jaw-teeth-face models in a variety of applications, including 3D digital treatment planning and orthognathic surgery. Difficulties in accurately merging facial scans and CBCT images are due to the different image acquisition methods and limited area of correspondence between the two facial surfaces. In addition, it is difficult to use machine learning techniques because they use face-related 3D medical data with radiation exposure, which are difficult to obtain for training. The proposed method addresses these problems by reusing an existing machine-learning-based 2D landmark detection algorithm in an open-source library and developing a novel mathematical algorithm that identifies paired 3D landmarks from knowledge of the corresponding 2D landmarks. A main contribution of this study is that the proposed method does not require annotated training data of facial landmarks because it uses a pre-trained facial landmark detection algorithm that is known to be robust and generalized to various 2D face image models. Note that this reduces a 3D landmark detection problem to a 2D problem of identifying the corresponding landmarks on two 2D projection images generated from two different projection angles. Here, the 3D landmarks for registration were selected from the sub-surfaces with the least geometric change under the CBCT and face scan environments. For the final fine-tuning of the registration, the Iterative Closest Point method was applied, which utilizes geometrical information around the 3D landmarks. The experimental results show that the proposed method achieved an averaged surface distance error of $0.74\,mm$ for three pairs of CBCT and face scan datasets.
\end{abstract}

}
\maketitle
\IEEEdisplaynontitleabstractindextext
\IEEEpeerreviewmaketitle

\begin{figure*}[ht]
	\centering
	\includegraphics[width=1\textwidth]{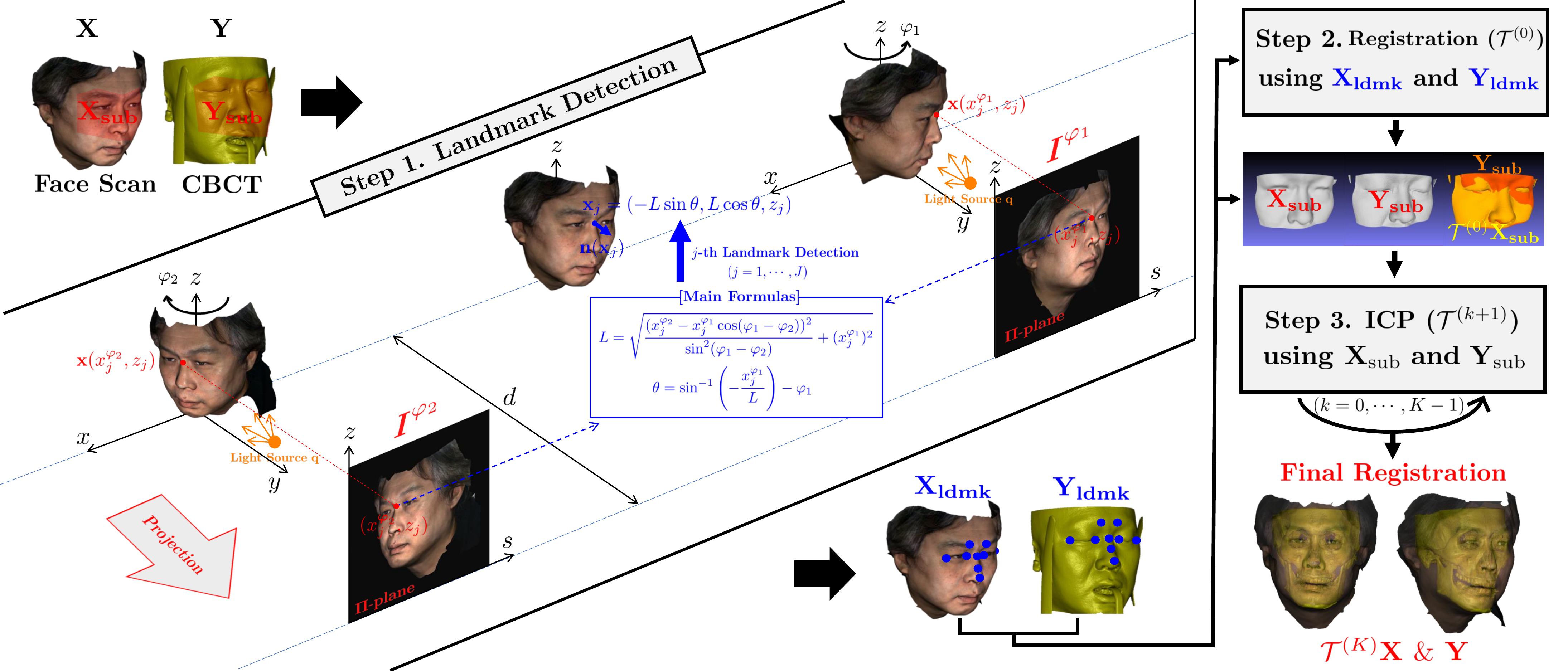}
	\caption{Schematic diagram of the proposed registration method of CBCT and face scan data. In the first step, the 3D facial landmarks $\X_{\mbox{\scriptsize ldmk}},\Y_{\mbox{\scriptsize ldmk}}$ on CBCT and face scan surfaces are detected using two different 2D projection images. In the second step, the rigid transformation ($\mT^{(0)}$) is obtained using the paired $\X_{\mbox{\scriptsize ldmk}}$ and $\Y_{\mbox{\scriptsize ldmk}}$. Finally, the ICP method is applied to accurately estimate $\mT$ using the sub-surfaces $\X_{\mbox{\scriptsize sub}}$ and $\Y_{\mbox{\scriptsize sub}}$.}
	\label{Schematic_diagram}
\end{figure*}

\section{Introduction}
With recent and rapid advances in dental imaging technologies, such as cone-beam computerized tomography (CBCT), intraoral scanners (IOS), and 3D facial scanners \cite{vandenberghe2020crucial}, there is a great need for the development of a smart digital platform of 3D jaw-teeth-face models to create an integrated patient treatment plan as a single digital anatomic model, including the bone, teeth, gingiva, and face \cite{elnagar2020digital,Shujaat2021}. 3D jaw-teeth-face models can be used to provide a prediction of surgical outcomes in patients with facial deformities and create simulations of various osteotomies with an idea of the expected esthetic changes \cite{elnagar2020digital,Joda2015}. It also enables dentists to explain treatment plans more effectively and facilitates communication between dentists and patients.

With the rapid enhancement of machine learning (ML) techniques, great success has been achieved in developing a fully automated method for integrating dental CBCT and IOS data into a jaw-tooth model that reaches the level of clinical application \cite{Chung2020,jang2021fully2}.

However, automatic integration between CBCT and face scans has not yet been achieved because of the dissimilarities between the data acquired from both devices \cite{Nahm2014,Schendel2009,Shujaat2021}. Dental CBCT and face scan data do not always contain complete point-to-point correspondences owing to the different acquisition environments. In dental CBCT scanning, unlike in the face scanning process, patients are asked to close their eyes to remove motion artifacts or bite plastic sticks to separate the upper and lower jaw bones in the CT image. In addition, dental CBCT has a limited field of view (FOV) that often does not fully cover the patient's head in the longitudinal direction. Another hurdle is the difficulty in using ML techniques, which have recently shown remarkable performance in registration \cite{Huang2021,Islam2021,Ma2021,Pais2020}. This is because it is difficult to collect paired training data of CBCT and the corresponding facial images, given the legal and ethical restrictions related to medical data.
This paper proposes a fully automatic CBCT-face scan registration method designed to address the aforementioned problems. The proposed method adopts a commonly used registration scheme that matches selected landmarks on both face surfaces, where the landmarks are selected from the portion of the face surface with the least geometric variation in the CBCT and face scan environments.

The main contribution of this study is the development of a new mathematical algorithm capable of converting a 3D landmark detection problem into a 2D problem of detecting the corresponding landmarks in two 2D projection images generated from different projection angles. This method allows the reuse of the existing ML method for 2D facial landmark detection provided in the open-source library Dlib \cite{King2009}. It is crucial to note that the proposed method does not require annotated training data of facial landmarks because it uses a pre-trained facial landmark detection algorithm using various public datasets. The Dlib facial landmark detection (DFLD) algorithm first detects a face in a 2D face image using a maximum-margin object detector with a convolutional neural network \cite{King2015} and then provides 68 different feature points on the face based on an Ensemble of Regression Trees \cite{Kazemi2014}. It is known to be robust and generalizable to various 2D face image models.

The proposed registration method is summarized as follows: the first step was to generate surfaces from the measured CT and facial scans. From each surface, two 2D projection images at different view angles were generated. We then detected the facial points corresponding to the 3D landmarks on the 2D projection images using the DFLD method. From the detected 2D facial points, the 3D landmark positions were estimated using a mathematical formula (See Theorem \ref{thm}). Using the multiple pairs of landmarks, the initial registration was performed. Finally, to improve the accuracy and efficiency of the registration of CT and face surfaces, the Iterative Closest Point (ICP) \cite{Arun1987} method was applied using sub-surfaces, including 3D landmarks. Detailed procedures are described in the Methods section.

\section{Method} \label{Method}
The goal of this study is to integrate 3D images from dental CBCT and facial scanners, which are different imaging modalities. Let $\X$ represent a 3D point cloud of a face surface obtained from a facial scanner. Let $\Y$ represent a 3D point cloud of the tissue surface obtained from a CBCT image. To achieve this goal, we align $\X$ and $\Y$ into a single coordinate system.
The 3D $\X-\Y$ registration requires choosing sub-surfaces $\X_{\mbox{\scriptsize sub}} \subseteq \X$ and $\Y_{\mbox{\scriptsize sub}}\subseteq \Y$ such that there exists a rigid transformation $\mathcal T : \X \rightarrow \Y$ satisfying
\begin{equation}\label{T}
	\mathcal T \X_{\mbox{\scriptsize sub}}  \approx \Y_{\mbox{\scriptsize sub}}.
\end{equation}
Here, the sub-surfaces $\X_{\mbox{\scriptsize sub}}$ and $\Y_{\mbox{\scriptsize sub}}$ should be chosen as the areas with the least geometric change in response to changes in the facial expression and movement of the mandible.
A rigid transformation $\mathcal T$ with six degrees of freedom can be determined if three or more independent landmarks are detected in $\X_{\mbox{\scriptsize sub}}$ and $\Y_{\mbox{\scriptsize sub}}$. Let $\X_{\mbox{\scriptsize ldmk}}= \{ \x_j = (x_j, y_j,z_j) : j=1,\cdots,J \} \subseteq \X_{\mbox{\scriptsize sub}}$ be landmark points and let $\Y_{\mbox{\scriptsize ldmk}}= \{  \y_j = (\hat x_j, \hat y_j,\hat z_j) : j = 1,\cdots,J \} \subseteq \Y_{\mbox{\scriptsize sub}}$ be the corresponding landmark points. The number $J$ must be $J \geq 3$. Given the pair of $\X_{\mbox{\scriptsize ldmk}}$ and $\Y_{\mbox{\scriptsize ldmk}}$, the rigid transformation $\mathcal T$ can be obtained by the following least-squares minimization:
\begin{align}\label{landmark_T}
 \mathcal T = \argmin_{\mathcal T}\sum_{j=1}^J\| \mathcal T \x_j-\y_j\|^2,
\end{align}
where $\|\cdot\|$ denotes the Euclidean norm.
Now, we describe a stable method for automatically detecting the 3D landmarks of $\X_{\mbox{\scriptsize ldmk}}$ and $\Y_{\mbox{\scriptsize ldmk}}$.

\subsection{Automatic 3D landmark detection using 2D projection images}
In this section, the automatic detection method of $\X_{\mbox{\scriptsize ldmk}}$ is explained because the detection of $\Y_{\mbox{\scriptsize ldmk}}$ is performed in the same manner. The direct detection of 3D landmarks $\X_{\mbox{\scriptsize ldmk}}$ on a 3D surface $\X$ with high confidence can be challenging. To address the difficulty of accurately detecting $\X_{\mbox{\scriptsize ldmk}}$ on $\X$, we transform the 3D detection problem into a relatively much easier 2D problem of detecting the corresponding landmark points on two 2D projection images generated from different projection angles, $\varphi_1$ and $\varphi_2$.

We now explain how to generate the projection image $\I^\varphi$ of the angle $\varphi$ mentioned above. We rotate the 3D face surface $\X$ around the $z$-axis by an angle of $\varphi$. Let $\X^\varphi$ represent the rotated surface. We select a plane of $\Pi=\{ (s, d,z): 1\le s, z\le 1024\}$, where $d$ is sufficiently large such that $\Pi$ lies outside of $\X$. The projection image $\I^\varphi$ is generated by the rotated surface $\X^\varphi$ onto the plane $\Pi$. See Figure \ref{Schematic_diagram} for $\Pi$ and $\I^\varphi$; to be precise, let $\x^\varphi (s,z)$ be a point in $\X^\varphi$ given by
\begin{equation}
	\x^\varphi(s,z) = \argmin_{\x \in \ell_{(s,z)}\cap\X^\varphi} \|\x - (s,d,z)\|,
\end{equation}
where $\ell_{(s,z)}$ is a line passing through $(s,d,z)$ and parallel to the $y$-axis. The image $\I^\varphi$ with a light source position at $\mathbf{q}$ is given by \cite{Lee2019,Lengyel2011}
\begin{align}
	\I^\varphi(s,z) = \max\left(\f{\n(\x^\varphi(s,z))\cdot(\mathbf{q}-\x^\varphi(s,z))}{\|\mathbf{q}-\x^\varphi(s,z)\|},0\right),
\end{align}
where $\n(\x)$ is the unit normal vector at $\x\in \X^\varphi$ .

Let $\{ (x^{\varphi}_j,z_j) : j = 1,\cdots, J \}$ denote 2D landmarks corresponding to 3D landmarks $\X_{\mbox{\scriptsize ldmk}}$. This is illustrated in Fig. \ref{Schematic_diagram}. Owing to the light source at $\mathbf{q}$, the projection image $\I^\varphi$ contains 3D geometric features that allow the detection of 2D landmarks using conventional techniques \cite{Dong2018,Kazemi2014,Wang2018,Wu2019}.

To detect 3D landmarks $\X_{\mbox{\scriptsize ldmk}}$, we generate the projection images $\I^{\varphi_1}$ and $\I^{\varphi_2}$ at two different angles, $\varphi_1$ and $\varphi_2$. From $\I^{\varphi_1}$ and $\I^{\varphi_2}$, we can easily obtain two sets of landmarks: $\{ (x^{\varphi_1}_j,z_j) : j =1, \cdots, J\}$ and $\{ (x^{\varphi_2}_j,z_j) : j =1, \cdots, J\}$.

\begin{figure}[t]
	\centering
	\includegraphics[width=0.9\textwidth]{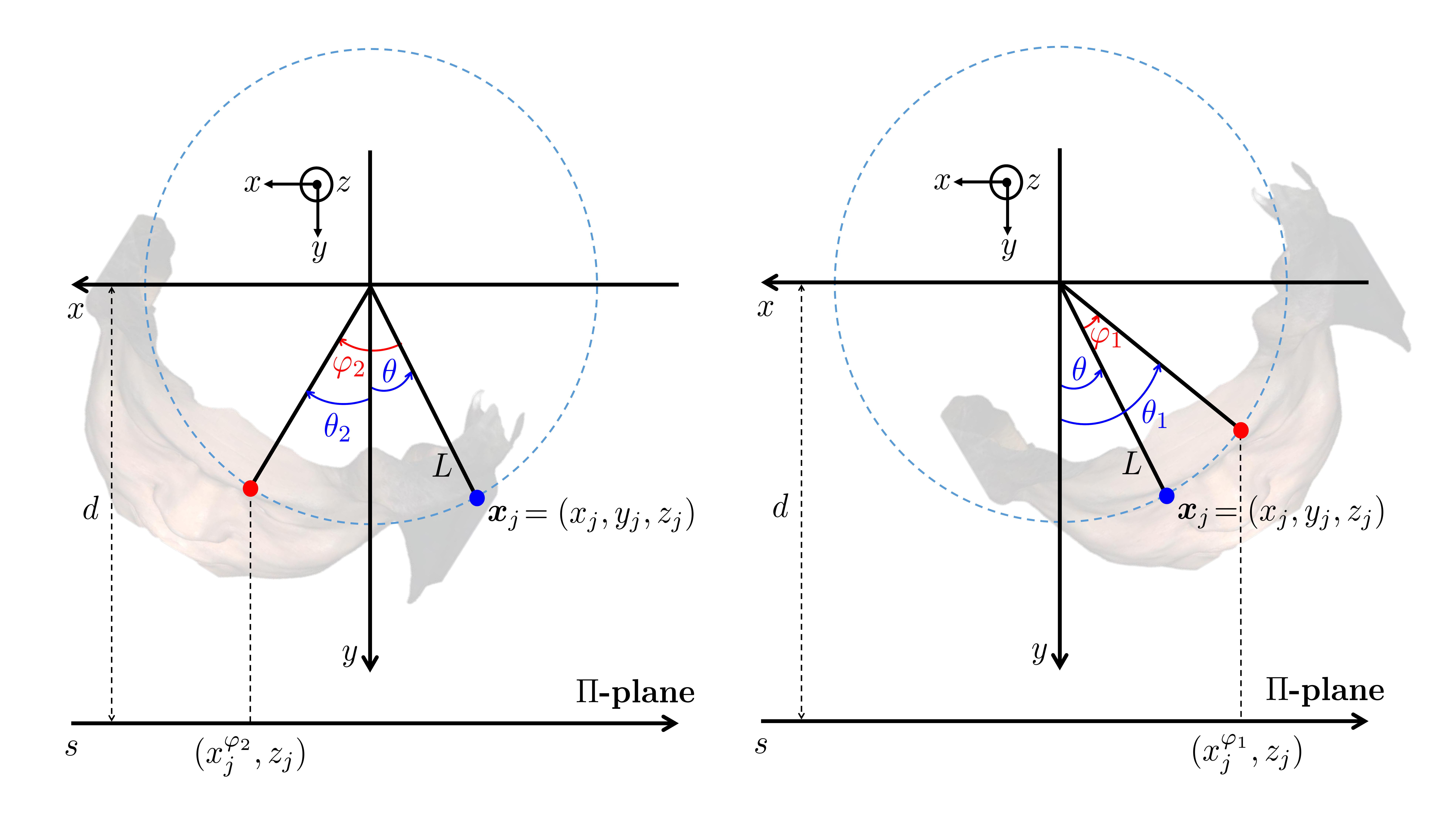}
	\caption{Geometries of 2D-3D landmarks}
	\label{Geometry}
\end{figure}

We now explain our method for detecting $\X_{\mbox{\scriptsize ldmk}}$. The proposed method uses the 2D landmarks $(x^{\varphi_1}_j,z_j)$ and $(x^{\varphi_2}_j,z_j)$ to identify the corresponding 3D landmark $\x_j=(x_j,y_j,z_j)$ through the following theorem.
\begin{theorem} \label{thm} Let two different rotation angles $\varphi_1$ and $\varphi_2$ be given. Suppose that the 2D landmarks $(x^{\varphi_1}_j,z_j)$ and $(x^{\varphi_2}_j,z_j)$, corresponding to the 3D landmark $\x_j$, are obtained from the two projection images $\I^{\varphi_1}$ and $\I^{\varphi_2}$, respectively. Let $\x_j$ be expressed by
\begin{align}\label{tri_formula}
	\x_j = (x_j,y_j,z_j) = (-L \sin \theta, L \cos \theta, z_j),
\end{align}
where $L=\sqrt{x_j^2 + y_j^2}$ and $\theta=\tan^{-1}(\dfrac{y_j}{x_j})-\dfrac{\pi}{2}$. Then, $L$ and $\theta$ are functions of $\varphi^1,\varphi^2,x^{\varphi_1}_j,$ and $x^{\varphi_2}_j$ that are respectively given by
\begin{align}\label{L-1}
	L & = L(\varphi^1,\varphi^2,x^{\varphi_1}_j,x^{\varphi_2}_j) \nonumber \\ & = \sqrt{\f{(x^{\varphi_2}_j-x^{\varphi_1}_j\cos(\varphi_1-\varphi_2))^2}{\sin^2(\varphi_1-\varphi_2)}+(x^{\varphi_1}_j)^2}
\end{align}
and
\begin{align}\label{theta}
	\theta = \theta(\varphi^1,\varphi^2,x^{\varphi_1}_j,x^{\varphi_2}_j) = \sin^{-1}\left(-\f{x^{\varphi_1}_j}{L}\right) - \varphi_1.
\end{align}
\end{theorem}
{\it \bf Proof}. Denoting $\theta_i\in[-\pi/2, \pi/2],~i=1,2$ by an angle from $\x^{\varphi_i}_{j}$ to the $yz$-plane, then $x^{\varphi_i}_{j}=-L\sin\theta_i,~i=1,2,$ $x_{j}=-L\sin(\theta_1-\varphi_1)$, and $y_{j}=L\cos(\theta_1-\varphi_1)$, where $L=\sqrt{x^2_{j} + y^2_{j}}$. See Fig. \ref{Geometry}. In addition, $\theta_i$ and $\varphi_i$ have the following relations:
\begin{align}\label{theta_relation}
\varphi_1 - \varphi_2 = \theta_1 - \theta_2.
\end{align}
It follows from the \eref{theta_relation} that the $x^{\varphi_2}_{j}$ is represented by:
\begin{align}\label{eq-1}
x^{\varphi_2}_{j} &= -L\sin(\theta_1 - (\varphi_1 - \varphi_2)) \nonumber \\
&= x^{\varphi_1}_{j}\cos(\varphi_1 - \varphi_2) + L\cos\theta_1\sin(\varphi_1 - \varphi_2).
\end{align}
It can be rewritten as
\begin{align}\label{eq-2}
L\cos\theta_1 = \f{x^{\varphi_2}_{j} - x^{\varphi_1}_{j}\cos(\varphi_1 - \varphi_2)}{\sin(\varphi_1 - \varphi_2)}.
\end{align}
Now, it follows from $L\sin\theta_1=-x^{\varphi_1}_{j}$ and \eref{eq-2} that
\begin{align}
L  = \sqrt{\f{(x_{j}^{\varphi_2}-x_{j}^{\varphi_1}\cos(\varphi_1-\varphi_2))^2}{\sin^2(\varphi_1-\varphi_2)}+(x_{j}^{\varphi_1})^2},
\end{align}
and
\begin{align}
\theta_1= \sin^{-1}\left(-\f{x^{\varphi_1}_{j}}{L}\right).
\end{align}
Denoting $\theta = \theta_1-\varphi_1$, this completes the proof.

\begin{figure}[h]
	\centering
	\includegraphics[width=1.0\textwidth]{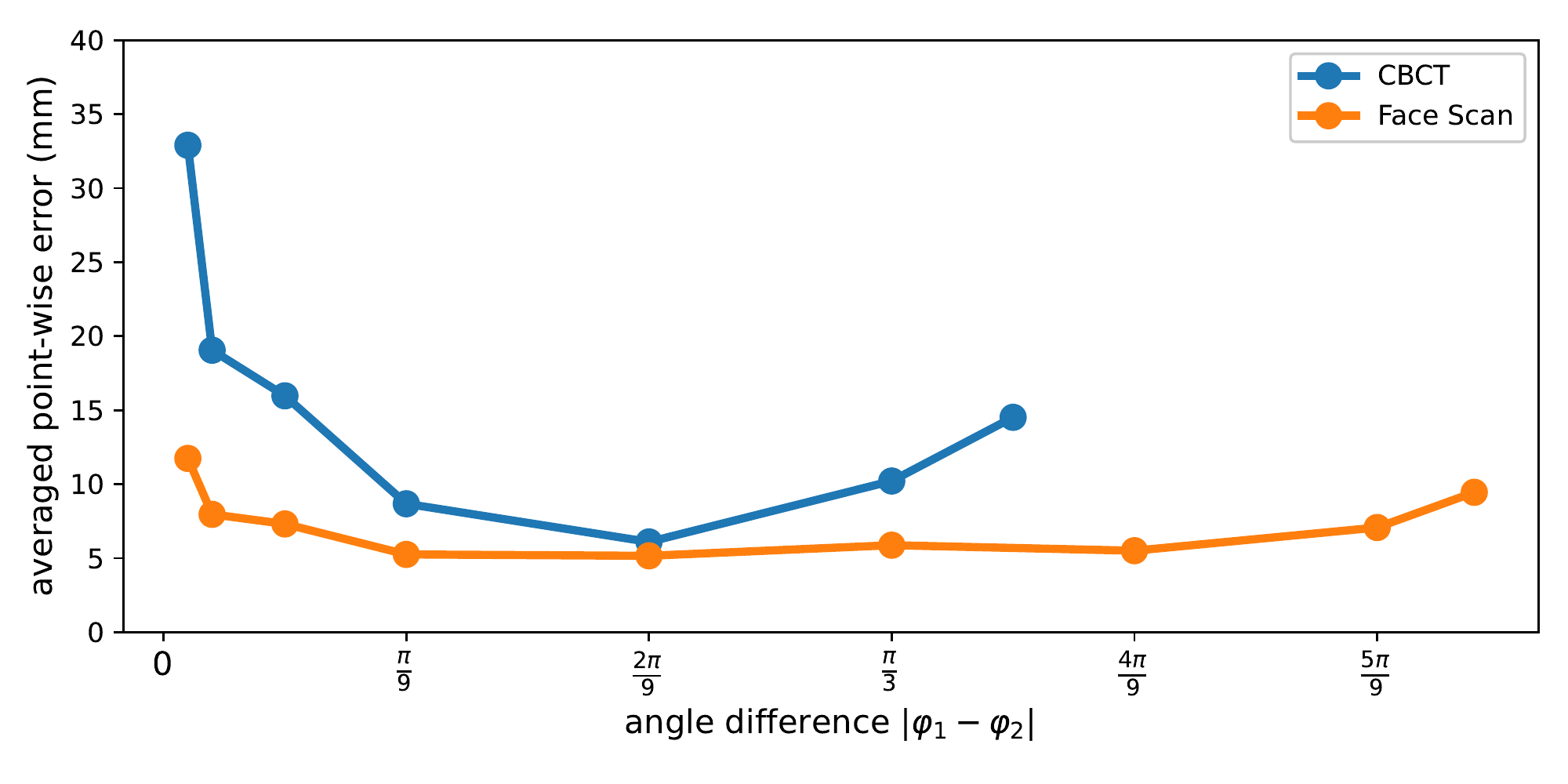}
	\caption{3D landmark detection error of CBCT and face scan with respect to the rotation angle.}
	\label{landmark_detection_errors}
\end{figure}

\begin{figure*}[ht]
	\centering
	\includegraphics[width=1\textwidth]{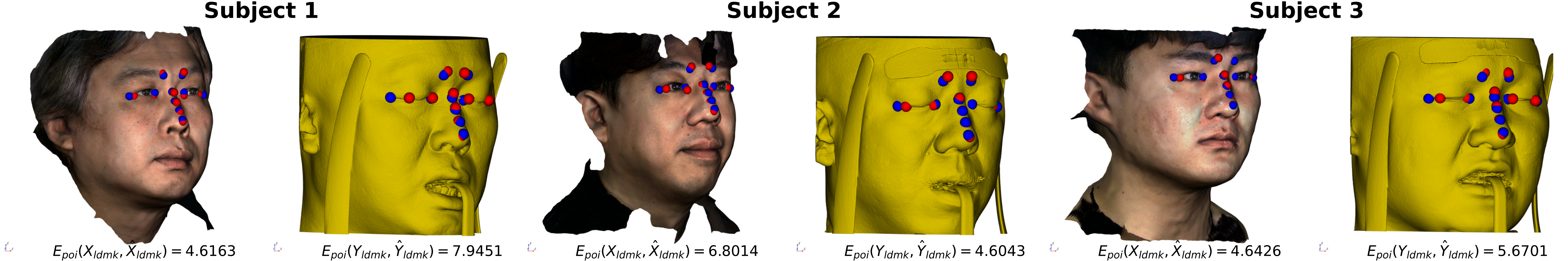}
	\caption{3D landmark detection results of CBCT and face scan surfaces. Blue balls represent the manually annotated landmark points. Red balls represent the points detected by the proposed algorithm. The point-wise error $E_{\text{poi}}$ between annotated and detected points are shown below each figure. }
	\label{landmark_detections_plot}
\end{figure*}

Note that the rotation angles $\varphi_1$ and $\varphi_2$ should be selected carefully. The difference $|\varphi_1 - \varphi_2|$ should not be too small or too large. Our experiments showed that either $|\varphi_1 - \varphi_2| \leq \frac{\pi}{9}$ or $|\varphi_1 - \varphi_2| \geq \frac{\pi}{3}$ tends to result in large errors in 3D landmark detection. This is illustrated in Fig. \ref{landmark_detection_errors}. The following corollary explains these errors theoretically.

\begin{corollary}\label{coro}
Let $x_j^{\varphi_1}$ and $x_j^{\varphi_2}$ be the detected 2D landmarks and let $\x_j$ be the estimated 3D landmark by the formulas \eqref{L-1} and \eqref{theta}. Assume that $\hat{\x}_j$,~$\hat{x}_j^{\varphi_1}$, and $\hat{x}_j^{\varphi_2}$ are the corresponding true landmark positions to $\x_j$, $x_j^{\varphi_1}$, and $x_j^{\varphi_2}$. Let $\hat{L}$ be the true value of $L$ in \eqref{tri_formula}. If the difference of projection angles, $|\varphi_1-\varphi_2| = \epsilon$, is small, the error $|L-\hat{L}|$ has the following asymptotic behavior as $\epsilon \rightarrow 0$:
\begin{align}\label{3d_error}
	|L-\hat{L}| ~ = ~ \epsilon^{-1} \left | |\hat{x}_j^{\varphi_2}-\hat{x}_j^{\varphi_1}|-|x_j^{\varphi_2}-x_j^{\varphi_1}| \right | + O(1),
\end{align}
where the symbol $O(\cdot)$ is the standard big-O asymptotic notation.
\end{corollary}
Note that the difference $|L-\hat{L}|$ in \eqref{3d_error} is closely related to the 3D detection error $\|\x_j-\hat{\x}_j\|$. The above corollary shows that $\|\x_j-\hat{\x}_j\|$ is magnified by a factor of $\epsilon^{-1}$.

{\it \bf Proof}. Without loss of generality, let $|\varphi_1 - \varphi_2| = \varphi_1 - \varphi_2$. Since $\epsilon \rightarrow 0$, the small angle approximation ($\sin \epsilon \approx \epsilon$ and $\cos \epsilon \approx 1$) can be applied to \eqref{L-1}, yielding the following relations:
\begin{align}
	L & = \sqrt{\f{(x^{\varphi_2}_j-x^{\varphi_1}_j\cos\epsilon)^2}{\sin^2\epsilon}+(x^{\varphi_1}_j)^2} \nonumber \\
	& = \sqrt{\epsilon^{-2}(x^{\varphi_2}_j-x^{\varphi_1}_j)^2+(x^{\varphi_1}_j)^2} + O(1) \nonumber \\
	& = \epsilon^{-1}|x^{\varphi_2}_j-x^{\varphi_1}_j| + O(1).
\end{align}
Hence, we obtain
\begin{equation} \label{Lerror}
	|L-\hat{L}| = \epsilon^{-1} \left||x^{\varphi_2}_j-x^{\varphi_1}_j| - |\hat{x}^{\varphi_2}_j-\hat{x}^{\varphi_1}_j| \right| + O(1).
\end{equation}
This completes the proof.

In the case when $|\varphi_1-\varphi_2| \geq \frac{\pi}{3}$, there is a high possibility that either $\langle \n(\x^{\varphi_1}_j(s,z)),(0,1,0) \rangle$ or $\langle \n(\x^{\varphi_2}_j(s,z)), (0,1,0) \rangle$ is very small, resulting in inaccurate detection of either $\x^{\varphi_1}$ or $\x^{\varphi_2}$.

\subsection{Fine registration using ICP method}
We adopt an ICP method to accurately estimate the optimal $\mathcal{T}$ in \eref{T} by using the information of $\X_{\mbox{\scriptsize sub}}$ and $\Y_{\mbox{\scriptsize sub}}$. Denoting $\mathcal{T}^{(0)}$ as the solution of \eref{landmark_T}, $\mathcal{T}$ is estimated by iteratively solving the following minimization problems \cite{Yang2015}. For $k=1,2,\ldots,K$,
\begin{align}\label{reg_eq1}
  \mT^{(k)} = \argmin_{\mathcal{T}} \f{1}{N}\sum_{i=1}^{N}  \|\mathcal{T}^{(k-1)}(\x_i)-\y_{j^*, \mT^{(k-1)}}\|^2,~\x_i\in\X_{\text{sub}}
\end{align}
where $\y_{j^*,\mT^{(k-1)}}$ is the closest point to the transformed point $\mT^{(k-1)}(\x_i)$. More precisely, for given $\mT^{(k-1)}$, $\y_{j^*,\mT^{(k-1)}}$ is given by
\begin{align}\label{reg_eq2}
  \y_{j^*,\mT^{(k-1)}} = \argmin_{\y_j\in\Y_{\text{sub}}} \|\mT^{(k-1)}(\x_i)-\y_j\|^2.
\end{align}
The detailed sub-surfaces $\X_{\mbox{\scriptsize sub}}$ and $\Y_{\mbox{\scriptsize sub}}$ are shown in Fig \ref{registration_results_distance}. These sub-surfaces can be estimated using the landmarks $\X_{\mbox{\scriptsize ldmk}}$ and $\Y_{\mbox{\scriptsize ldmk}}$.

\subsection{Datasets}
Three dental CBCT scans were acquired using a commercial CBCT scanner (RAYSCAN Studio, Ray Co., Ltd.) with a tube voltage of 90 kVp and tube current of 8 mA. The 3D CBCT images of size $666\times 666\times 666$ were reconstructed with a pixel size of $0.3\times0.3\,mm^2$ and slice thickness of $0.3\,mm$. The FOV of CBCT was $20 \text{ cm}\times 20 \text{ cm}$. From the reconstructed CT images, a surface (i.e., soft tissue) was extracted with a threshold value of $-500\text{ HU}$. Then, the surface mesh was generated using the standard matching-cube algorithm \cite{Lorensen1987}, where a point cloud of the surface model was obtained. Corresponding surface data from the facial scans were acquired using a commercial facial scanner (RayFace 200, Ray Co., Ltd.).

We collected a total of 20 multi-detector CT (MDCT) scans and their corresponding face scans for further evaluation. The MDCT scans were obtained using a commercial MDCT scanner (SOMATOM Force, Siemens) with a tube voltage of 100 kVp and a current of 120 mA. The MDCT slice images of size $512\times512$ were reconstructed with a pixel size of $0.46\times0.46\,mm^2$. The number and thickness of slices varied depending on acquisition conditions. The average number and thickness of slices were $530\,mm$ and $0.55\,mm$, respectively. To visually match the MDCT images to CBCT images, the MDCT images of the upper cranium were cropped. The surface data from MDCT scans was generated in the same manner as for the CBCT scans, using a commercial 3D scanner (EinScan Pro, Shining 3D).

\subsection{Evaluation metrics for 3D landmark detection and surface registration}
To quantitatively measure the 3D landmark detection accuracy, we calculated the point-wise error between the landmark $\mathbf{K}_{\text{ldmk}}$ and the corresponding true landmark $\hat{\mathbf{K}}_{\text{ldmk}}$,  $\mathbf{K} = \X,\Y$, by
\begin{align}\label{point_error}
  E_{\text{poi}}(\mathbf{K}_{\mbox{\scriptsize ldmk}},\hat{\mathbf{K}}_{\mbox{\scriptsize ldmk}})  = \sqrt{ \f{1}{J}\sum_{j=1}^J \|\mathbf{k}_{j}-\hat{\mathbf{k}}_{j}\|^2},
\end{align}
where $\mathbf{k}_{j}$ and $\hat{\mathbf{k}}_{j}$ are elements of $\mathbf{K}_{\mbox{\scriptsize ldmk}}$ and $\hat{\mathbf{K}}_{\mbox{\scriptsize ldmk}}$, respectively. In our study, $\hat{\mathbf{K}}_{\mbox{\scriptsize ldmk}}$ was obtained manually. To quantitatively measure the registration accuracy, we also computed the surface error between the two sub-surfaces $\mT(\X_{\mbox{\scriptsize sub}})$ and $\Y_{\mbox{\scriptsize sub}}$ using the following two metrics \cite{Rockafellar2009}:
\begin{align}\label{surf_error}
  &E^{\text{sup}}_{\text{surf}}(\mT(\X_{\text{sub}}),\Y_{\text{sub}}) = \sup_{\x\in\mT(\X_{\text{sub}})}\inf_{\y\in\Y_{\text{sub}}}\|\x-\y\|
\end{align}
and
\begin{align}\label{surf_error_ave}
  &E^{\text{mean}}_{\text{surf}}(\mT(\X_{\text{sub}}),\Y_{\text{sub}}) = \frac{1}{\vert\X_{\text{sub}}\vert}\sum_{\x\in\mT(\X_{\text{sub}})}\inf_{\y\in\Y_{\text{sub}}}\|\x-\y\|,
\end{align}
where sup, inf, and $\vert\X_{\text{sub}}\vert$ denote the supremum, infimum, and the number of points in the point cloud set $\X_{\text{sub}}$, respectively. Here, $E^{\text{ave}}_{\text{surf}}$ and $E^{\text{sup}}$ represent the mean and maximum surface errors over $\mT(\X_{\text{sub}})$, respectively.

\subsection{Implementation details}
We adopted a pre-trained DFLD model to detect the 2D landmarks $\{ (x^{\varphi}_j,z_j)\}_{j=1}^J$ and $\{ (y^{\varphi}_j,z_j)\}_{j=1}^J$ corresponding to the 3D landmarks $\X_{\text{ldmk}}$ and $\Y_{\text{ldmk}}$, respectively. The DFLD model is known to be robust and computationally efficient in the computer vision field because it is learned using various public datasets, including ImageNet, VOC, and VGG. The DFLD model can detect 68 2D face landmarks related to the eyebrows, eyes, nose, lips, and face contours in 2D images. From the 68 positions, we empirically selected 10 positions with the least positional change in the surfaces of the CBCT and face scans and regarded them as our facial landmarks $\X_{\text{ldmk}}$ and $\Y_{\text{ldmk}}$. These positions are shown in detail in Fig. \ref{Schematic_diagram}.

We evaluated the 3D landmark detection error with respect to the rotation angle $\varphi$. Throughout this study, we set $\varphi_2=-\varphi_1,~\varphi_1>0$. The averaged point-wise errors $E_{\text{poi}}(\X_{\mbox{\scriptsize ldmk}},\hat{\mathbf{X}}_{\mbox{\scriptsize ldmk}})$ and $E_{\text{poi}}(\Y_{\mbox{\scriptsize ldmk}},\hat{\mathbf{Y}}_{\mbox{\scriptsize ldmk}})$ for three subjects are shown in Fig. \ref{landmark_detection_errors}. The error increased when $|\varphi_1-\varphi_2|$ either decreased or increased. The overall error for face scan was lower than that for CBCT. However, the minimum errors for the CBCT and face scans were comparable. The DFLD method failed to detect $x^{\varphi_1}$ in the 2D projection image of CBCT for $|\varphi_1-\varphi_2| > 7\pi/18$.  In our study, we selected $|\varphi_1-\varphi_2|=2\pi/9$ with the minimum error for 3D landmark detection.

\begin{figure*}[ht]
	\centering
	\includegraphics[width=1\textwidth]{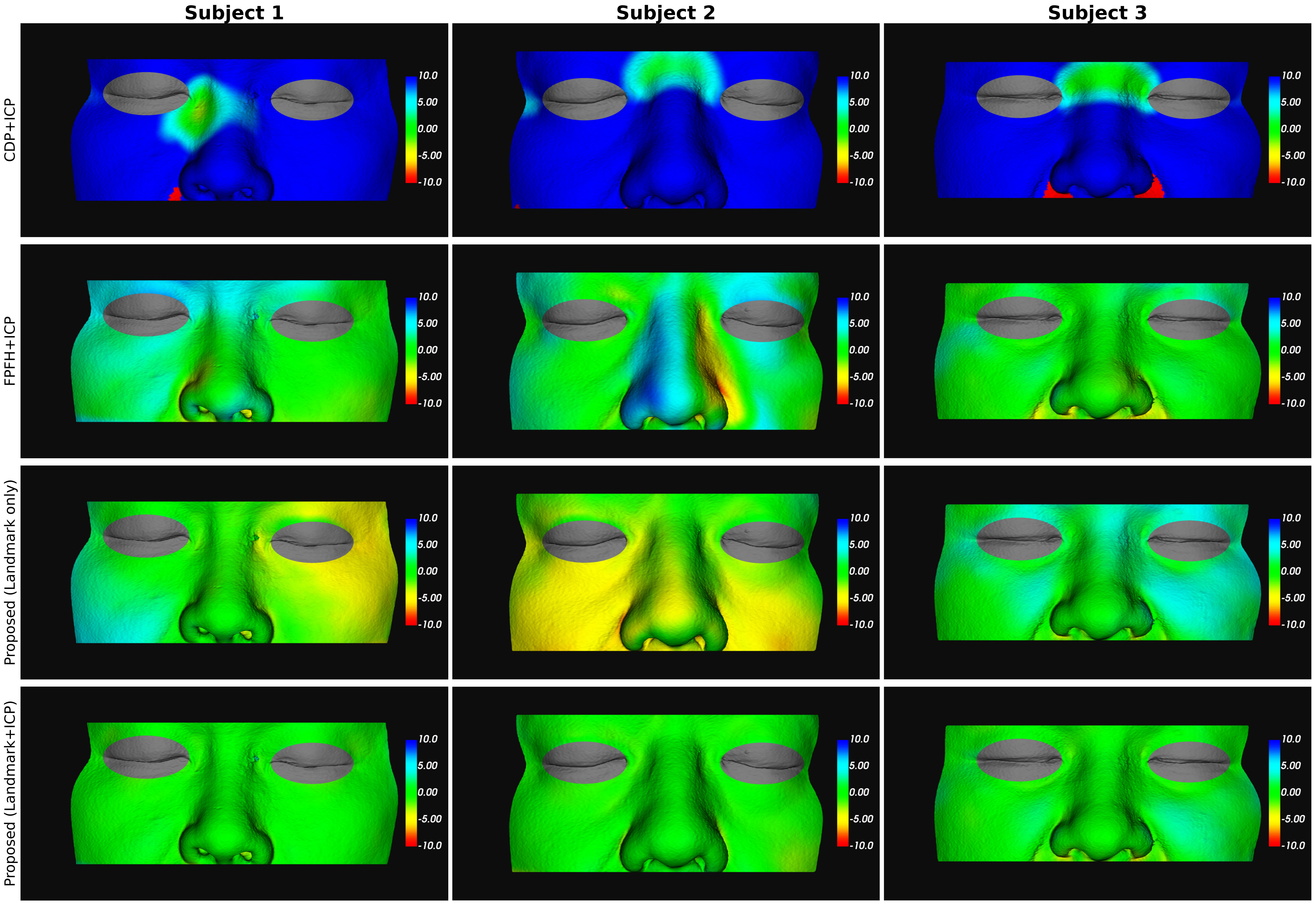}
	\caption{Comparison results of the registration methods for three subjects. The colors represent distances between the two surfaces obtained by CBCT and face scan. The unit is $mm$.}
	\label{registration_results_distance}
\end{figure*}

\begin{figure*}[!]
	\centering
	\includegraphics[width=1\textwidth]{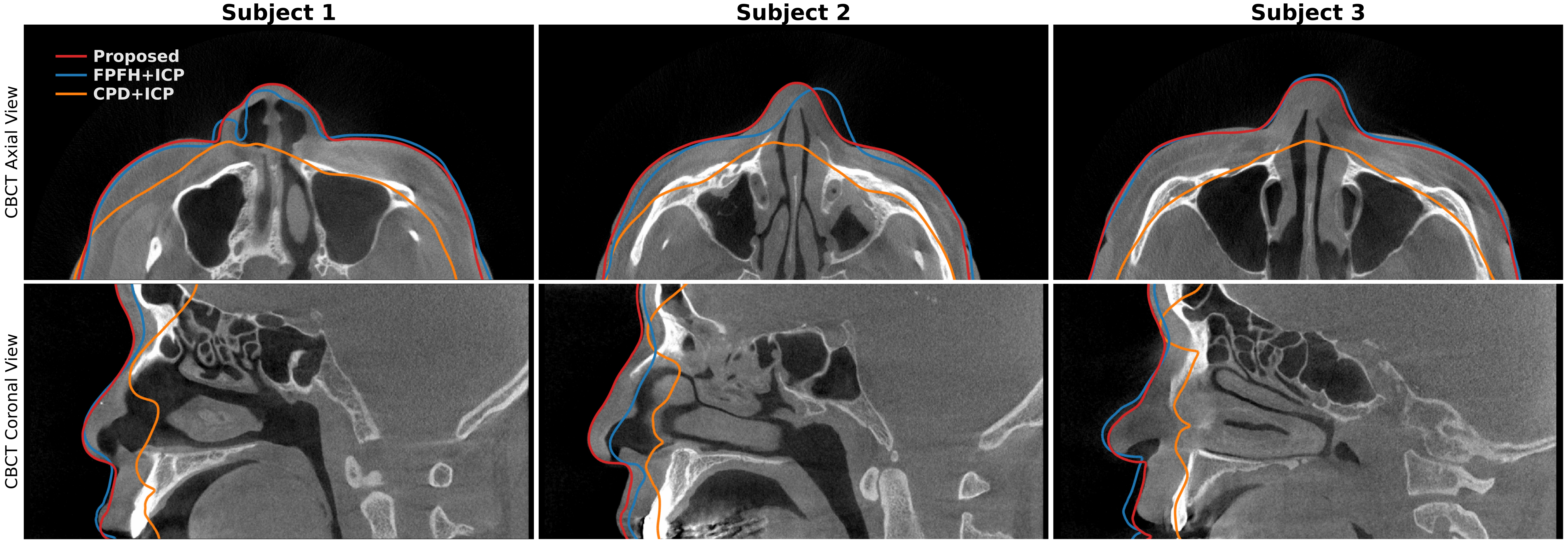}
	\caption{Comparison results of the registration methods for three subjects. The results are visualized on CBCT images. Red, blue, and orange solid lines represent the face scan surfaces aligned with the proposed method (landmark+ICP) and the FPFH and CDP followed by the ICP, respectively.}
	\label{registration_results_sectionplane}
\end{figure*}

\begin{table*}
\centering
\begin{tabular}{llcccc}\toprule
\textbf{Metric}   &  \textbf{Method}  &  \multicolumn{4}{c}{\textbf{Subject}}  \\
\cmidrule(r){3-6} &                   & 1             & 2            & 3  & Mean $\pm$ Std \\
\midrule
\multirow{4}{*}{$E^{\text{sup}}_{\text{surf}}(\mT(\X_{\text{sub}}),\Y_{\text{sub}})$}
                 & CDP+ICP           			&  35.8175      & 44.4622      & 44.3512  		&  41.5436$\pm$4.0492    \\
                 & FPFH+ICP          			&  10.6258      & 7.8111       & 6.3459   		&  8.2609$\pm$1.7759 \\
                 & Proposed (landmark only)   	&  6.6144       & 8.2409       & 7.8157   		&  7.5570$\pm$0.6887  \\
                 & Proposed (landmark+ICP)      & {\bf 3.6862}  & {\bf 3.4213} & {\bf 5.3387}  	& {\bf  4.1487$\pm$0.8483 }   \\
\midrule
\multirow{4}{*}{$E^{\text{mean}}_{\text{surf}}(\mT(\X_{\text{sub}}),\Y_{\text{sub}})$}
                 & CDP+ICP           			&  7.5566    	& 16.1463      & 14.5461  		&  12.7496$\pm$3.7297 \\
                 & FPFH+ICP          			&  2.0766    	& 1.9783       & 1.3639   		&  1.8062$\pm$0.3153 \\
                 & Proposed (landmark only)   	&  2.2389    	& 3.6479       & 1.8230   		&  2.5699$\pm$0.7809 \\
                 & Proposed (landmark+ICP)      & {\bf 0.5093}  & {\bf 0.7563} & {\bf 0.9487}   & {\bf 0.7381$\pm$0.1798 }   \\
\bottomrule
\end{tabular}
\caption{Quantitative evaluation of the registration methods for CBCT and face scan datasets. The surface errors $E^{\text{sup}}_{\text{surf}}$ and $E^{\text{mean}}_{\text{surf}}$ were computed for three subjects. The unit is $mm$.}
\label{table_quantitative_registration}
\end{table*}

\begin{table*}
\centering
\begin{tabular}{llc}\toprule
\textbf{Metric}   &  \textbf{Method}  & Mean $\pm$ Std \\
\midrule
\multirow{4}{*}{$E^{\text{sup}}_{\text{surf}}(\mT(\X_{\text{sub}}),\Y_{\text{sub}})$}
                 & Manual           			&  2.2863 $\pm$ 0.6965    \\
                 & CDP+ICP           			&  30.3500 $\pm$ 17.3744    \\
                 & FPFH+ICP          			&  3.1801 $\pm$ 1.2598 \\
                 & Proposed (landmark only)   	&  7.6889 $\pm$ 3.5975  \\
                 & Proposed (landmark+ICP)      &  {\bf 2.2818 $\pm$ 0.7098} \\
\midrule
\multirow{4}{*}{$E^{\text{mean}}_{\text{surf}}(\mT(\X_{\text{sub}}),\Y_{\text{sub}})$}
                 & Manual           			&  0.4227 $\pm$ 0.1174    \\
                 & CDP+ICP           			&  9.0067 $\pm$ 5.8236 \\
                 & FPFH+ICP          			&  0.6997 $\pm$ 0.4568 \\
                 & Proposed (landmark only)   	&  3.1006 $\pm$ 2.1206 \\
                 & Proposed (landmark+ICP)      & {\bf 0.3749 $\pm$ 0.1289}   \\
\bottomrule
\end{tabular}
\caption{Quantitative evaluation of the registration methods for MDCT and face scan datasets. The surface errors $E^{\text{sup}}_{\text{surf}}$ and $E^{\text{mean}}_{\text{surf}}$ were computed for ten subjects. The unit is $mm$.}
\label{table_quantitative_registration_mdct}
\end{table*}

\section{Results} \label{Results}

Fig.~\ref{landmark_detections_plot} presents the 3D landmark detection results of the proposed method regarding the surfaces of the CBCT and face scans. The proposed method stably detected 3D landmarks on both surfaces for all three subjects. The point-wise errors $E_{\text{poi}}(\X_{\mbox{\scriptsize ldmk}},\hat{\mathbf{X}}_{\mbox{\scriptsize ldmk}})$ and $E_{\text{poi}}(\Y_{\mbox{\scriptsize ldmk}},\hat{\mathbf{Y}}_{\mbox{\scriptsize ldmk}})$ for the three subjects were computed. The computed values are shown below each panel in Fig.~\ref{landmark_detections_plot}. The average errors of the CBCT and face scans for the three subjects were $6.0733\,mm$ and $5.3534\,mm$, respectively.

We compared the registration performance of the proposed method with those of existing global registration methods, such as Coherent Point Drift~(CPD)~\cite{Myronenko2010} and Fast Point Feature Histograms (FPFHs)~\cite{Rush2009}, followed by those of ICP methods. Fig.~\ref{registration_results_distance} shows the registration results for the sub-surfaces $\Y_{\text{sub}}$ of the three subjects. In Fig.~\ref{registration_results_distance}, the color at each point represents the signed distance between the two surfaces obtained by CBCT and face scanning. More precisely, for each $\y\in \Y_{\text{sub}}$, the signed distance $\mathfrak{d}(\y)$ is calculated as
\begin{align}
  \mathfrak{d}(\y)=\text{sgn} ((\x^*-\y)\cdot\n(\y))\|\x^*-\y\|,
\end{align}
where sgn$(\cdot)$ is a sign function and $\x^*\in \mT(\X_{\text{sub}})$ is a point satisfies
\begin{align}
  \|\x^*-\y\|=\inf_{\x\in\mT(\X_{\text{sub}})} \|\x-\y\|.
\end{align}
Overall, the proposed method aligned the two surfaces more accurately than existing registration methods. The results also imply that the proposed method, using only 3D landmarks (third row in Fig. \ref{registration_results_distance}), provides a good initial transformation for the ICP compared with these existing global registration methods. The corresponding registration results were compared visually in the CBCT images, as shown in Fig. \ref{registration_results_sectionplane}.

For a quantitative comparison, we computed the surface errors $E^{\text{sup}}_{\text{surf}}(\mT(\X_{\text{sub}}),\Y_{\text{sub}})$ and $E^{\text{mean}}_{\text{surf}}(\mT(\X_{\text{sub}}),\Y_{\text{sub}})$ of the registration methods for the three subjects, where the detailed sub-surfaces of $\X_{\text{sub}}$ and $\Y_{\text{sub}}$ are shown in Fig. \ref{registration_results_sectionplane}. The computed errors are presented in Table~\ref{table_quantitative_registration}. Among the registration methods, the proposed method achieved the minimum mean surface errors of $4.1487\,mm$ and $0.7381\,mm$ for the metrics $E^{\text{sup}}_{\text{surf}}$ and $E^{\text{mean}}_{\text{surf}}$, respectively.

Table \ref{table_quantitative_registration_mdct} presents the quantitative results of the registration methods for MDCT and face scan datasets. In the case of the MDCT dataset, the performance of the proposed registration method was additionally compared to that of manual registration by an expert in oral and maxillofacial surgery, who had more than 20 years of experience, followed by ICP. The proposed method outperformed both the global registration methods and manual registration, achieving the lowest mean surface errors for the metrics $E^{\text{sup}}{\text{surf}}$ and $E^{\text{mean}}{\text{surf}}$, with values of $2.2818\,mm$ and $0.3749\,mm$, respectively.

\section{Discussion and Conclusion} \label{Conclude}
This paper proposes a fully automatic registration method between dental CBCT and facial scan data. Noting that the facial surface obtained from the facial scanner corresponded only partially to that obtained from dental CBCT, the proposed method was designed to match the portion of the facial surface with the smallest geometrical change using a 3D geometric landmark registration approach. The novel mathematical formulation described in Theorem \ref{thm} can reduce a 3D landmark detection problem to a 2D problem of detecting the corresponding landmarks on two 2D projection images generated from two different projection angles. This reduction allows robust detection of 3D landmarks by leveraging a pre-trained 2D facial landmark detection algorithm. A major advantage of reusing a pre-trained 2D landmark detection algorithm is that the cumbersome and costly problem of collecting the annotations of facial landmarks for training is eliminated. Experiments demonstrated that the proposed method outperformed other existing global registration methods, such as CPD and FPFH, followed by ICP. The proposed method achieved a mean surface error of $0.7381\,mm$ and $0.3749\,mm$ for CBCT and MDCT cases, respectively. In particular, in the case of MDCT, it was attained similar or lower mean errors compared with manual registration results of expert, which makes it possible to expect the capability to use in clinical application. 

The proposed landmark-based registration method can be applied to dental CBCT with a large FOV, covering the patient's nose to eyebrows. Recently, numerous commercial products with large FOVs (e.g., $16\,cm \times 16\,cm$, $16\,cm \times 23\,cm$, $20\,cm\times 20\,cm$) have been released for visualizing the entire craniofacial area, and they can be beneficial for orthodontics, airway studies, and oral surgery \cite{Agrawal2013,Kapila2015}. A future research aim is to develop a fully automatic method of non-rigid registration \cite{Bahri2021} between CBCT and facial scans to match the surface areas near the mouth where large geometric deformations occur. The machine learning can be used to effectively represent non-rigid transformations.

\section*{Acknowledgments}
This work was supported by a grant of the Korea Health Technology R$\&$D Project through the Korea Health Industry Development Institute (KHIDI), funded by the Ministry of Health $\&$ Welfare, Republic of Korea  (HI20C0127).
H.S.P. and K.J. were partially supported by  the National Institute for Mathematical Sciences (NIMS) grant funded by the Korean government (No. NIMS-B23910000). C.M.H and J.K.S were partially supported by Samsung Science $\&$ Technology Foundation (No. SRFC-IT1902-09).

\end{document}